\documentclass[conference]{IEEEtran}
\usepackage[utf8]{inputenc}
\IEEEoverridecommandlockouts
\usepackage{cite}
\usepackage[hyphenbreaks]{breakurl}
\usepackage{amsmath,amssymb,amsfonts}
\usepackage{algorithmic}
\usepackage{graphicx}
\usepackage{textcomp}
\usepackage{xcolor}

\usepackage{blindtext}
\usepackage{subfigure}
\usepackage{diagbox}
\usepackage{booktabs}
\usepackage{threeparttable}

\def\BibTeX{{\rm B\kern-.05em{\sc i\kern-.025em b}\kern-.08em
    T\kern-.1667em\lower.7ex\hbox{E}\kern-.125emX}}
    
\graphicspath{{figures/}}
\begin{document}

\title{INTERNEURON: A Middleware with Multi-Network Communication Reliability for Infrastructure Vehicle Cooperative Autonomous Driving}

\author{Tianze Wu, 
Shaoshan Liu,
Bo Yu,
Sa Wang,
Yungang Bao,
Weisong Shi
}
\if false
\author{\IEEEauthorblockN{Tianze Wu\IEEEauthorrefmark{1}\IEEEauthorrefmark{3}, 
Shaoshan Liu\IEEEauthorrefmark{2},
Bo Yu\IEEEauthorrefmark{2},
Sa Wang\IEEEauthorrefmark{3},
Yungang Bao\IEEEauthorrefmark{3},
Weisong Shi\IEEEauthorrefmark{2}
}
\IEEEauthorblockA{\IEEEauthorrefmark{1}SKL of Computer Architecture, Institute of Computing Technology, CAS, Beijing, China}
\IEEEauthorblockA{\IEEEauthorrefmark{2}University of Delaware, USA}
\IEEEauthorblockA{\IEEEauthorrefmark{3}University of Chinese Academy of Sciences, Beijing, China}
{\{wutianze, wangsa, baoyungang\}}@ict.ac.cn, {\{Weisong\}}@udel.edu}
\date{March 2022}
\fi
\maketitle

\begin{abstract}
Infrastructure-Vehicle Cooperative Autonomous Driving (IVCAD) is a new paradigm of autonomous driving, which relies on the cooperation between intelligent roads and autonomous vehicles. This paradigm has been shown to be safer and more efficient compared to the on-vehicle-only autonomous driving paradigm.  Our real-world deployment data indicates that the effectiveness of IVCAD is constrained by reliability and performance of commercial communication networks.  This paper targets this exact problem, and proposes INTERNEURON, a middleware to achieve high communication reliability between intelligent roads and autonomous vehicles, in the context of IVCAD. Specifically, INTERNEURON dynamically matches IVCAD applications and the underlying communication technologies based on varying communication performance and quality needs. Evaluation results confirm that INTERNEURON reduces deadline violations by more than 95\%, significantly improving the reliability of IVCAD systems.
\end{abstract}


\section{\textbf{introduction}}
\label{introduction}

In the past decade, on-vehicle-only autonomous driving has made a promise to revolutionize our transportation systems \cite{liu2020creating}. However, the promise has yet to be delivered and we are still far away from ubiquitous deployment of fully autonomous vehicles. 

If we delve into the problem of autonomous driving progress stagnation, each autonomous vehicle has only limited sensing and computing capability, hence it is challenging for autonomous vehicles to gather comprehensive information from their environment, and they are not intelligent enough to operate on their own with only limited information\cite{liu2019edge}. Infrastructure Vehicle Cooperative Autonomous Driving (IVCAD), on the other hand, utilizes infrastructure-side sensing and computing capabilities to empower all autonomous vehicles to be smarter, and furthermore make the whole transportation system more efficient \cite{liu2022road}.  

On the technical details \cite{liu2021towards}, a IVCAD system consists of the System-on-Vehicles (SoVs), or the autonomous driving system installed on each vehicle; the System-on-Roads (SoRs), or the autonomous driving system installed on road side; and intelligent transportation cloud system (ITCS), or the central cloud system to coordinate the autonomous vehicles to achieve maximum traffic efficiency.

During operation of an IVCAD system, the SoRs provide local perception results to the SoVs for blind spot elimination and extended perception to improve safety. The ITCS fuses all incoming data to generate global perception and planning information, and plans navigation trajectories for each vehicle to achieve optimal traffic efficiency \cite{liu2022brief}.

While IVCAD has been proven to be an effective autonomous driving paradigm, in our real-world IVCAD deployments, we have identified network communication as the main technical roadblock for reliable cooperative autonomous driving \cite{liu2022wcm}: the current mobile network bandwidth is constraining the uploading of raw sensing data, which is crucial for cloud-based applications such as deep learning model training and high-definition map generation \textit{etc}. Second, the network latency jitters remain high for a considerable portion of a vehicle's trip, greatly impacting the reliability and safety of the operations of autonomous vehicles. One snapshot of the seriousness of the network communication problem can be found in Table \ref{tab:4g_5g}, which shows an example of latency variation in 4G and 5G communication networks. These heavy variations greatly impact reliable deployments of IVCAD applications. 

\begin{table}[h]
\caption{Real-World Network Latency}
\label{tab:4g_5g}
\begin{threeparttable}
\begin{tabular}{ccc}
\hline
              & 4G Mobile Network & 5G Mobile Network \\ \hline
Average (ms)  & 173               & 26                \\
Min. (ms)     & 23                & 19                \\
Max. (ms)     & 994               & 96                \\
Standard Dev. (ms) & 258               & 12                \\ \hline
\end{tabular}
\begin{tablenotes}
\footnotesize
\item[*] The measurement session is conducted along our test field route on our commercial deployment site with an average speed of 50 km/h.
\end{tablenotes}
\end{threeparttable}
\end{table}

This paper targets this exact problem, and proposes \emph{INTERNEURON}, a middleware to achieve optimal communication reliability in the context of IVCAD, under real-world multi-network environments. Note that although \emph{INTERNEURON} is software only, evaluation results have verified that \emph{INTERNEURON} effectively reduces deadline violations by more than 95\%. In summary, this paper makes the following contributions:

\begin{itemize}
\item We propose \emph{INTERNEURON}, a middleware to achieve optimal communication reliability in the context of IVCAD. Based on the IVCAD application need, \emph{INTERNEURON} dynamically allocates the communication workload to the most suitable underlying network. Section \ref{design} provides the details of the \emph{INTERNEURON} design.
\item We propose a real-time guarantee mechanism in \emph{INTERNEURON} for not just IVCAD but all cooperative robots. The proposed mechanism combines network, data and computation tradeoff with multi-network environment to extend the runtime of distributed real-time systems as much as possible. Detailed discussion on this can also be found in Section \ref{design}.
\item We have evaluated the effectiveness of \emph{INTERNEURON} and demonstrate that \emph{INTERNEURON} reduces deadline violations by more than 95\%, hence significantly improving the reliability of IVCAD systems. More details on the evaluation methods and results can be found in Section \ref{evaluation}. 
\end{itemize}

\section{\textbf{background}}
\label{background}

In this section, we review the technologies related to \emph{INTERNEURON} in the context of IVCAD, including computation offloading, networking challenges, and communication middleware. 

\subsection{Computation Offloading}

As many autonomous driving workloads are extremely compute-intensive, offloading to edge and cloud has been proven an effective mitigation technique \cite{liu2017computer}. A key problem facing computation offloading is the exchange of external computational resources for minimal communication overhead. Along this line, Neurosurgeon\cite{kang2017neurosurgeon} splits DNNs on mobile devices and data centers with neural network layers granularity to find the best partition points of different models through a large amount of profiling. Kayak \cite{ you2021ship} combines two methods, ship compute and ship data, making full use of the computational resources of both application server and storage server. \cite{jiang2021joint} demonstrates that by combining different input data and model versions can obtain similar latency and accuracy metrics, and then A2 is proposed to adjust data accuracy and model versions to meet latency and accuracy requirements. Data compression, on the other hand, is an effective way to reduce communication overheads.  \cite{du2020server,yao2020deep}, \textit{etc}. 
propose different compression techniques. Based on various optimization techniques proposed in prior art, \emph{INTERNEURON} focuses on integrating these optimization techniques to provide real-time guarantees, which are essential to autonomous driving.

\subsection{Networking Challenges in IVCAD}

Networking has been the main challenge for IVCAD since the inception of IVCAD \cite{liu2022wcm,liu2021towards}. First, there is a contradiction between the low reliability of wireless network communication and the high reliability required by autonomous driving. Even with the latest 5G technology, there are still many network fluctuations reported that prevent 5G commercial networks to be reliably utilized for IVCAD. Second, the uncertainty of mobile networks \cite{lucia_et_al:LIPIcs:2017:7131} leads to the difficulty of establishing real-time assurance mechanisms for IVCAD. Real-time systems require a highly deterministic system environment, so they can only be implemented on a single machine. It remains a challenge to establish a real-time system in a distributed computing environment \cite{choi2016making}, where the problem of time synchronization of different nodes alone is very difficult to solve. Without a reliable real-time mechanism, IVCAD will only be a simple aid to autonomous driving. Third, mobile network technology evolves rapidly \cite{ahangar2021survey}, leading to many communication technologies with varying standards used in IVCAD \cite{san2020precise,parthasarathy2016vehicle,fitah2018performance,molina2017lte}. We believe that the underlying communication technologies will continue to iterate rapidly for the foreseeable future, and that the co-existence of multiple communication technologies will be a distinctive feature of IVCAD. \emph{INTERNEURON} targets these exact challenges and provides solutions to the above three problems that IVCAD currently faces.

\subsection{Communication Middleware}

Middleware is the management layer between OS and the upper applications\cite{schantz2002middleware}, with the target to provide common services and capabilities such as task scheduling, inter-process communication, data management outside of what is offered by OS\cite{redhatmiddle}. Existing autonomous driving solutions \cite{kato2015open,apex.org,apollo.com} tend to run on a single machine, so the middleware layer plays a limited role in optimizing the communication performance. In contrast, IVCAD is a highly distributed system suffering from high and varying communication latency\cite{wu2021oops}, and because most current IVCAD applications only need to transmit simple state information, the existing middleware does not optimize the cross-node communication deeply. Existing middleware such as FastDDS\cite{fastrtps} does not provide QoS guarantee. Recently, \cite{gog2022d3} proposes ERDOS, which guarantees real-time performance by dynamically adjusting the deadlines of computational tasks according to the current environment. However, ERDOS focuses on single machine deployments and only considers computation time while ignoring communication time. Inspired by ERDOS, \emph{INTERNEURON} endows cross-node communication and data processing with the ability to dynamically adjust run time, ultimately providing end-to-end (e2e) latency guarantees.
\section{Design of INTERNEURON}
\label{design}
\begin{figure}[htb]
\centering
\includegraphics[width=1\linewidth]{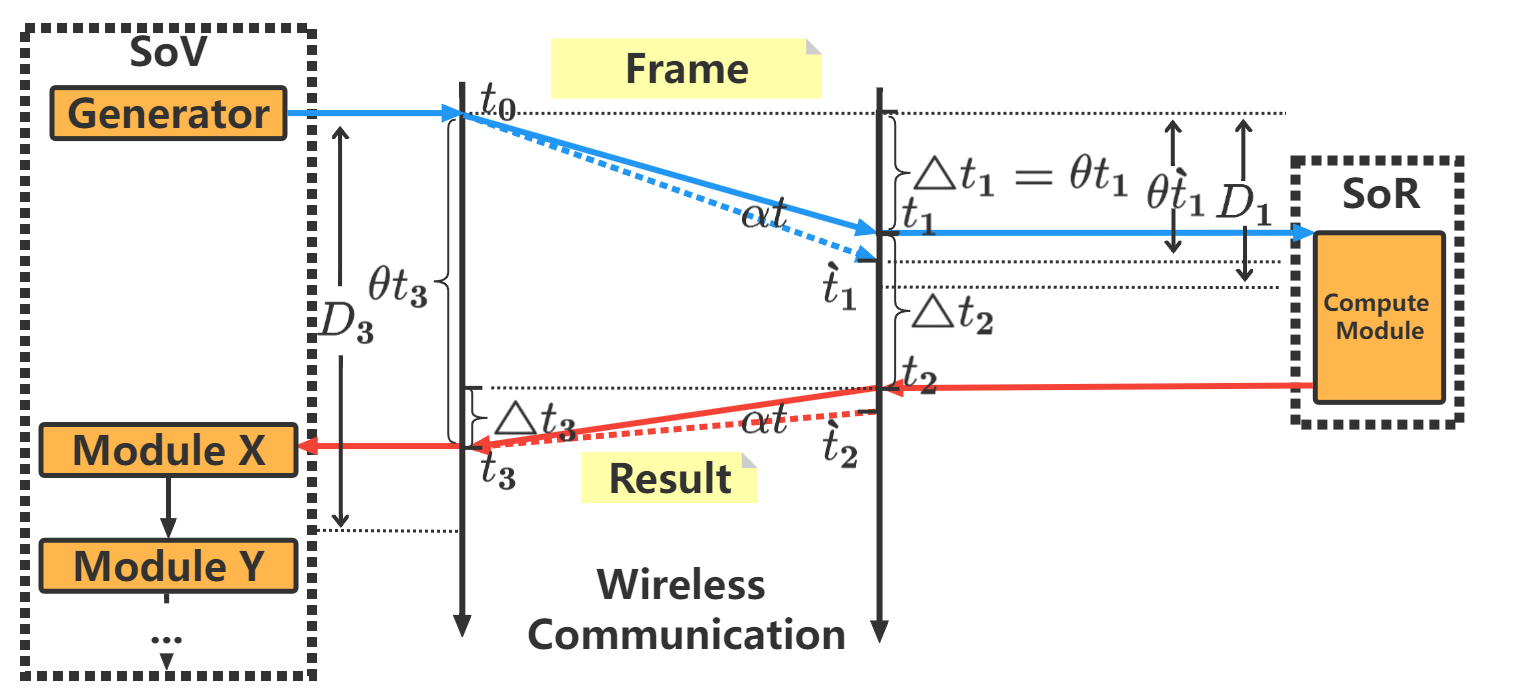}
\caption{\textbf{Typical Pipeline.} A frame generated at time $t_0$ marks the beginning of the pipeline. The frame is then sent through wireless network to SoR. It is received at $t_1$ and the transmission delay is $\triangle t_1 = t_1 - t_0$. Since the two different devices are not synchronized, the actual timestamps $\grave t_1$ \& $\grave t_2$ recorded by SoR are $\alpha t$ different from the time of SoV. After being handled by the Compute Module in SoR, the result is sent back to SoV at $t_2$ and be finally received at $t_3$. We denote the time taken by each computation or transmission task as $\triangle t_n$, the elapsed time from $t_0$ till current time point $t_n$ as $\theta t_n$, and the timeout moment of each $\theta t_n$ as $D_n$. In particular, the e2e delay from generating the frame to getting the processing result is $D_3$.}
\label{fig:typical_pipeline}
\end{figure}

This section first discusses the design requirements for the middleware of IVCAD systems. Then we describe the architectural and technical details of our proposed middleware system. To better explain our design, we take the typical pipeline in Fig.~\ref{fig:typical_pipeline} as an example, where the symbols used in this paper are also defined.

\subsection{Design Requirements}
In this subsection, we list the design requirements as follows:
\textbf{R1:} an unified abstraction and interface for applications transparent to the underling V2X hardware, and provide applications with the ability to specify communication QoS requirements. \textbf{R2:} awareness of changing network environments and capable of adaptive optimization in runtime. \textbf{R3:} real-time guarantee under varying network conditions. \textbf{R4:} reliable executions tolerant to network performance jitters.

\begin{figure}[htb]
\centering
\includegraphics[width=1\linewidth]{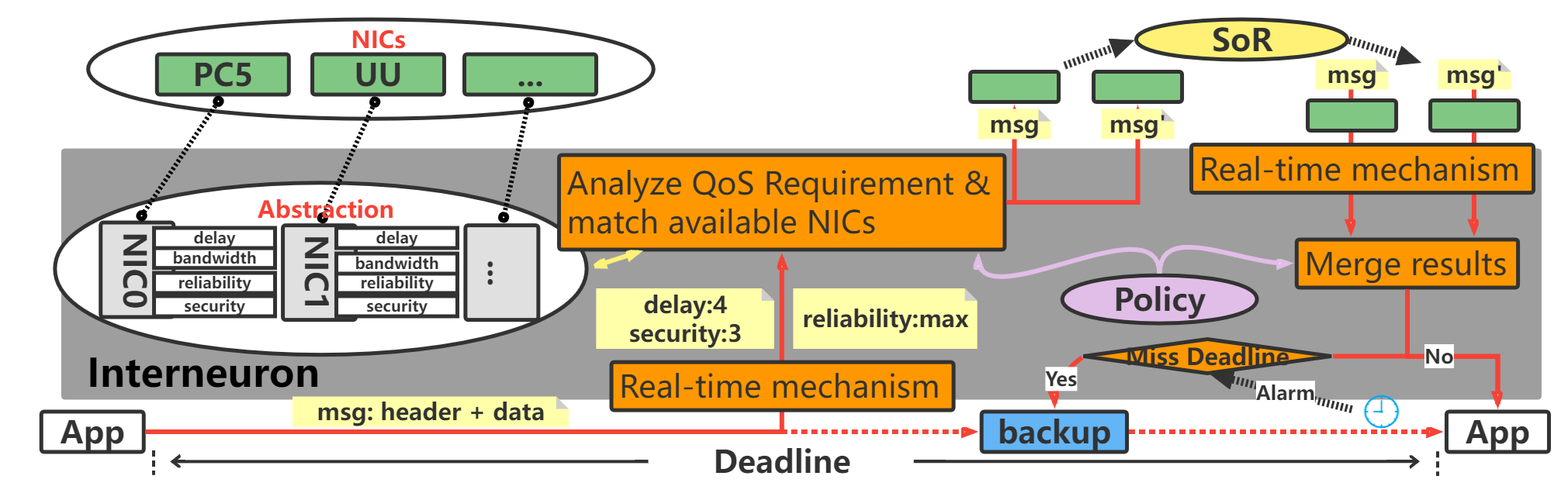}
\caption{\textbf{Architecture of INTERNEURON.} \emph{INTERNEURON} is located between applications and underlying network devices. By abstracting the Network Interface Cards (NICs) on the machine and evaluating their performance (recorded as \emph{Nature}), \emph{INTERNEURON} is able to match suitable NICs to complete transmission according to the \emph{policy} and QoS requirements. Similarly, upon receiving results from SoR, \emph{INTERNEURON} would merge results from different NICs according to the \emph{policy}. A local backup is also started at the beginning, and its results will be used if the results from SoR is not ready by the deadline.}
\label{fig:arch}
\end{figure}

\subsection{Architecture of INTERNEURON}
As shown in Fig.~\ref{fig:arch}, \emph{INTERNEURON} is located between applications and network devices. It has two main responsibilities: one is to provide \textit{QoS guaranteed network communication services} for upper-layer applications using multi-network resources; the other is to dynamically \textit{guarantee the real-time performance} of the entire pipeline according to the current running state. \emph{Policy} in the figure decides how to utilize multi-network resources to meet the QoS requirements when sending messages. Also, \emph{policy} is responsible for merging messages received from different Network Interface Cards (NICs). The task of the real-time mechanism is to analyze the current running state, and then propose QoS requirements for the next communication task (\emph{policy}'s responsibility) or computation task (application's responsibility) based on the analysis results. Last but not least, a local backup task is started on the beginning of the pipeline, it is able to keep the system functioning when the e2e deadline is missed.

\subsection{Network Abstraction}
To achieve R1 and R2, a software abstraction layer is required the hide the details of the underlying network technologies. As shown in Fig.~\ref{fig:arch}, \emph{policy} only needs to care about how to utilize the currently available communication resources to meet the QoS requirements, the change of underlying devices should have no effect on the \emph{policy}'s logic. Specifically, \emph{INTERNEURON} uses four fields named delay, bandwidth, reliability, security to describe the "\emph{Nature}" of one network. \emph{Nature} basically cover the communication requirements of the application. A larger value of a field indicates that the corresponding NIC performs better on this feature than other NICs. For example, NIC A scores higher in "delay" than NIC B, which means that the message delivery delay through A tends to be lower than B. 

\subsection{Communication \emph{Policy}}
The network abstraction layer effectively decouples \emph{policy} and network devices. If the \emph{policy} needs to use network functions, it only needs to provide a \emph{Nature} requirement table for \emph{INTERNEURON} to automatically find the network devices that meet the requirements. The requirement table is calculated by the \emph{policy} based on the QoS requirements of the application (placed in message header) and real-time mechanism. For example, if an application needs to transmit a piece of sensitive and private information, then its QoS requirement will be extremely security demanding. Then, the \emph{policy} logic may decide to use the NIC with highest security field score. Since \emph{policy} does not depend on underlying devices or upper-layer applications, developers could easily customize \emph{policy} to adapt to other scenarios (R2). 

\textbf{Down-Sampling} This \emph{policy} is targeted at sensor messaging scenarios. As we know, most of the sensor data processing takes place locally in current IVCAD system, because the stable transmission of big data is very demanding for the network. The most effective way to relieve the network pressure is to reduce the message size. There are many ways to do this job, and we use down-sampling as an example. If the message to be sent is a image, down-sampling processing can change the picture from RGB to black and white, or take samples at intervals of several pixels; If the message is point cloud data, then down-sampling processing can reduce the density of point cloud data. The down-sampled data will be much smaller than the raw data. If the application has a QoS requirement on the delay of transmission, this policy will select two networks with the highest delay score from the currently available networks that meet the basic bandwidth requirement for image transmission. Under normal conditions, the network with less bandwidth will be responsible for transmitting the down-sampled data, while the network with more bandwidth will transmit the raw data. Both networks transmit down-sampled data only when the real-time mechanism specifically requires it.

On the receiver side, if the down-sampled data arrives first, the \emph{policy} temporarily stores the data and waits for real-time mechanism's command. If the real-time mechanism decides not to wait for the raw data to arrive, the \emph{policy} would directly return the down-sampled data to the application.

\subsection{Real-Time Guarantee}

To satisfy R3, we have developed a real-time guarantee mechanism for \emph{INTERNEURON}. As shown in Fig.~\ref{fig:arch}, the real-time guarantee mechanism is located at the entrance of \emph{INTERNEURON}. Whenever \emph{INTERNEURON} receives a message from upper-layer App or underlying network, the real-time mechanism will analyze the running state and guide the following operations. Next, we introduce the details of our design, starting with the challenges that distributed environment poses to real-time guarantee.

\textbf{Challenges in a distributed environment:} The main difference between a distributed environment and a single-node environment is the additional overhead of cross-node communication between modules in a distributed environment. This poses two fatal problems for real-time guarantee mechanisms: 1. Time is not synchronized between different devices. While there is a lot of work exploring clock synchronization mechanisms\cite{geng2018exploiting, li2020sundial, liu2021brief}, these mechanisms usually rely on stable network environment and are expensive, making it difficult for them to work in an IVCAD environment. 2. The uncertainty in communication delay is large. For computation tasks on a single node, the jitters of run time can be reduced by means of operating system or even hardware, such as resource isolation. However, the delay of cross-node communication tasks is difficult to avoid jitters. \cite{choi2016making} has made some attempts in SDN environment, but it is still unable to do anything in the face of unstable wireless environment like IVCAD. Time asynchronism combined with uncertain communication delay leads to the system not even being able to determine whether this current moment has timed out.

Our real-time mechanism addresses these challenges in the following ways:

\textbf{The elapsed time of this round is compared with a reference time to avoid the problem of time asynchronism:} Take the scenario in Fig.~\ref{fig:typical_pipeline} as an example, because of the time difference $\alpha t$, SoR considers the current elapsed time to be $\theta \grave t_1 = \grave t_1 - t_0$, while the actual elapsed time is $\theta t_1 = t_1 - t_0 = \grave t_1 - \alpha t - t_0$. Since it is hard to get $\alpha t$ in IVCAD environment, we choose to introduce a reference elapsed time $\theta \grave t^{base}_0$ to avoid $\alpha t$. We assume that $\theta \grave t^{base}_0$ is the value recorded by SoR in some previous perfect run of the pipeline. Since that run was in the most perfect state, we could evaluate the state of the current run by comparing $\theta \grave t_1$ and $\theta \grave t^{base}_1$ as follows: $\theta \grave t_1 - \theta \grave t^{base}_1 = \theta t_1 + \alpha t - \theta t^{base}_1 - \alpha t = \theta t_1 - \theta t^{base}_1$. In this way, we managed to avoid the influence of $\alpha t$.

\begin{figure}[htb]
\centering
\includegraphics[width=1\linewidth]{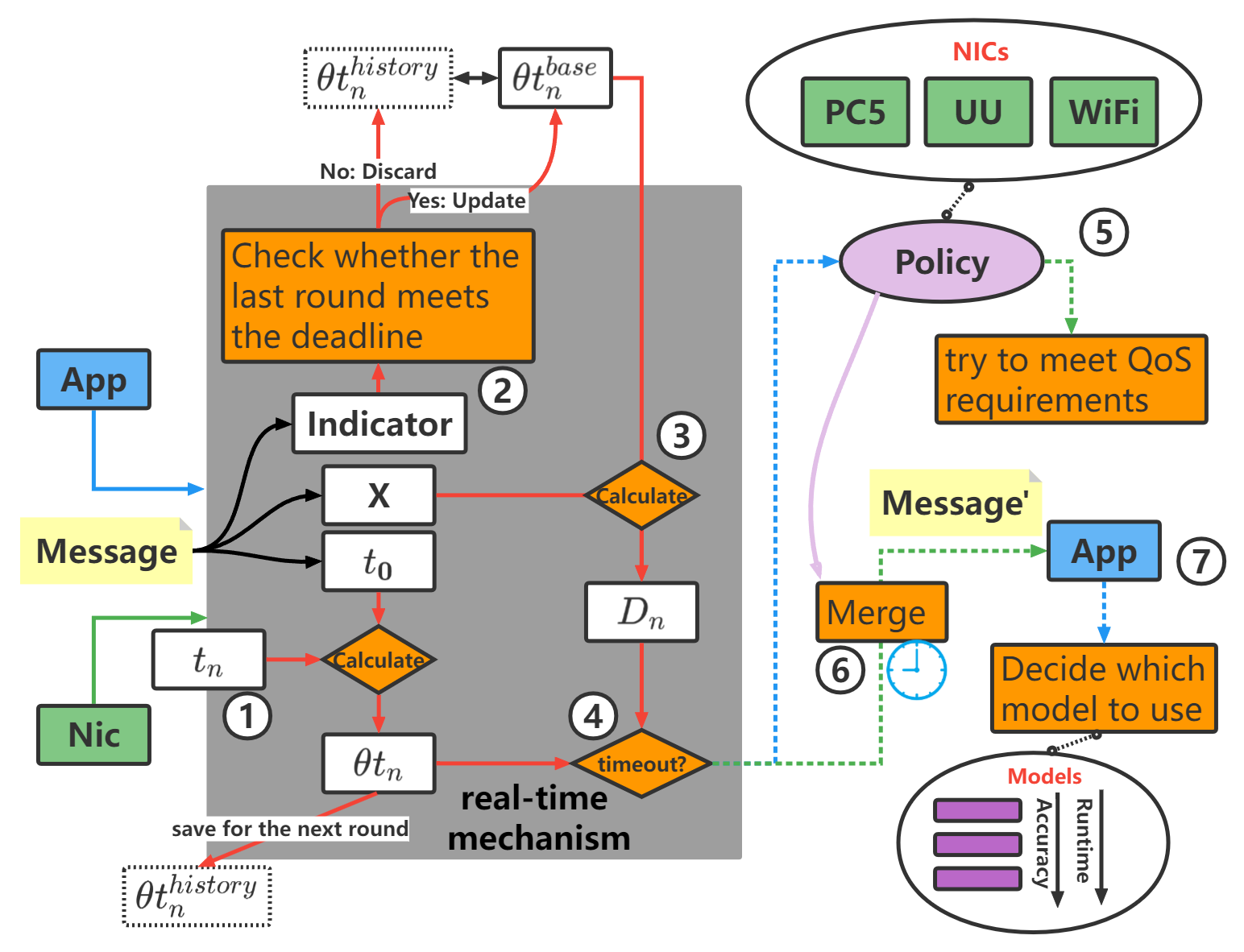}
\caption{\textbf{Real-time mechanism.} Upon receiving a message, a timestamp $t_n$ is recorded, then \emph{INTERNEURON} calculates the time already elapsed $\theta t_n = t_n-t_0$ (\textbf{$t_0$} is carried by the message header) and save it for next round (1). Whether the previous round meets the e2e deadline is stored in \textbf{Indicator} of the message header, which decides whether to discard the previous $\theta t^{history}_n$ or use it to update the $\theta t^{base}_n$ (2). The upper limit of $\theta t_n$ ($D_n$) is calculated using \textbf{X} (carried by the message header) and $\theta t^{base}_n$ (3). By comparing $\theta t_n$ and $D_n$ (4), \emph{INTERNEURON} could tell whether a timeout happens. This information affects the policy for subsequent calculations or message delivery. If the message comes from application layer, \emph{INTERNEURON} would utilize multi-network resources to meet QoS requirements according to a \emph{policy} (5). If the message comes from underlying NICs, \emph{INTERNEURON} would first merge messages from different NICs into one piece (Message') according to a \emph{policy} (6). The header of Message' contains the info whether the application needs to take actions to reduce its run time. Then, the application could decide which model to use according to the header of Message' (7).}
\label{fig:rt_mech}
\end{figure}

\textbf{Obtain the reference time by collecting recent performance statistics online:} Then, the problem is how we could provide a "perfect" reference time $\theta \grave t^{base}_1$? On a single machine, the reference time for a computation module can be obtained by offline profiling. However, as mentioned earlier, the communication latency is very unstable in the wireless network environment of IVCAD, so online profiling is almost the only option. Through testing, we find that the communication latency jitter of a certain network in IVCAD environment is not completely disordered. Take 4G, 5G and other mobile networks for example, most of the time, their performance is relatively stable. The problem occurs mainly in the following scenarios: 1. When the vehicle is driving away from the communication base station; 2. At the junction of multiple base station coverage areas; 3. A large number of nearby devices cause signal interference or network congestion. Delays in recent periods could reflect the current network status accurately. Therefore, we choose to use recent $\theta t_n$ data to calculate $\theta t^{base}_n$ (here and later, we use $\theta t_n$ instead of $\theta \grave t_n$ for convenience). The data being used must meet the requirement that the pipeline on which it is running does not end up with an e2e timeout. It should be noted that the reference time mechanism can be used not only for communication tasks, but also for computation tasks.

Implementation details are shown in Fig.~\ref{fig:rt_mech}. $\theta t^{base}_n$ is the reference time we need, and $\theta t^{history}_n$ is the $\theta t_n$ of the previous round which is saved in step (1). \textbf{Indicator} in the message header tells the real-time mechanism whether the last round meets the e2e deadline. If not, $\theta t^{history}_n$ does not meet the requirement and will be discarded. Otherwise, $\theta t^{base}_n$ would be updated as follows: $\theta t^{base}_n = \theta t^{base}_n * k + \theta t^{history}_n * (1-k), k \in [0,1]$. \textbf{k} is used to smooth the change curve of the reference time. It should be noted that we do not need to record a $\theta t^{base}_n$ for each NIC individually; we only care about how much time the transmission module cost.

\textbf{Determine the current running state:} With a "perfect" reference time, the next step is how to judge the current running state based on it. Specifically, if $\theta t_n < \theta t^{base}_n$ which means previous tasks are completed ahead of schedule, it is obvious that current state is good. However, if $\theta t_n > \theta t^{base}_n$, it is hard to decide whether the current state is poor. Consider the following scenario, a recent e2e run time of the pipeline has been much less than the specified timeout. Thus, even if $\theta t_n$ is a little larger than $\theta t^{base}_n$, it may not cause serious consequences. So we define a threshold ratio \textbf{X} as shown in Fig.~\ref{fig:rt_mech}, only when $(\theta t_n - \theta t^{base}_n)/\theta t^{base}_n$ exceeds X that we say the current state is bad. We define $D_n = X * \theta t^{base}_n$, $D_n$ as the upper limit of $\theta t_n$ that the real-time mechanism can tolerate. To reflect the above run-time surplus scenario, we let $X = 1 - \theta t^{base}_3/D_3 + \beta$, where $\theta t^{base}_3$ is the e2e reference run time of the previous round and $D_3$ is the preset deadline of the pipeline demonstrated in Fig.~\ref{fig:typical_pipeline}. The regulation amount $\beta$ is used for special adjustments. 

\begin{figure}[htb]
\centering
\includegraphics[width=1\linewidth]{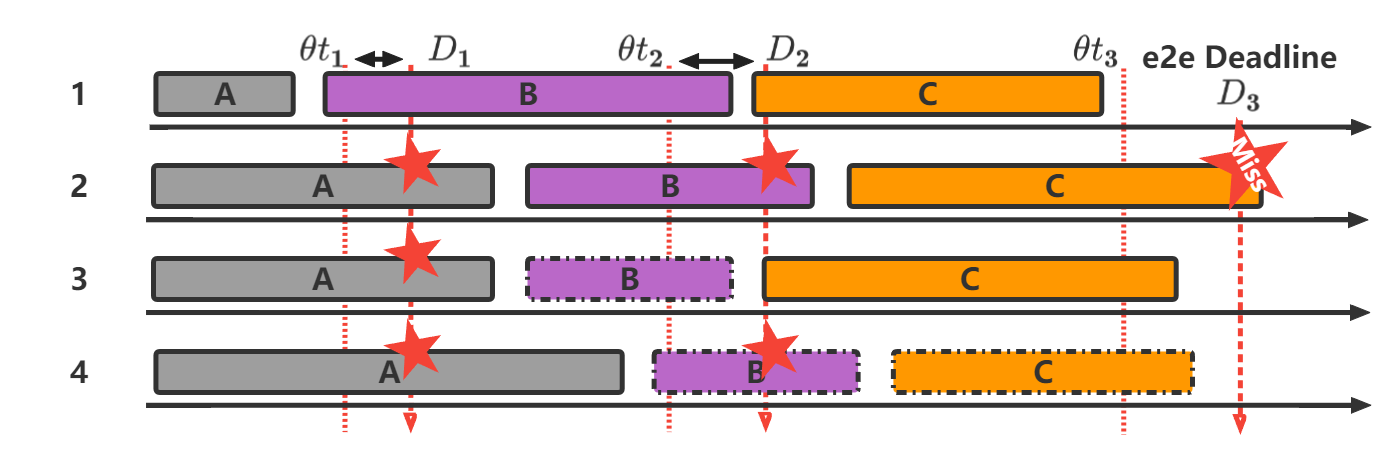}
\caption{\textbf{Real-time guarantee in pipeline.} A, B, C are three modules (Computation or Communication Module) in a pipeline. Line 1 is a normal round of the pipeline, each module finishes without missing the deadline, and the e2e deadline would not be missed obviously. In Line 2, the delay of A increases and the first deadline $D_1$ is missed. If no action is taken, even B and C cost the same time as Line \emph{base}, the e2e deadline will be missed. Line 3 demonstrates how we deal with the problem. When \emph{INTERNEURON} detects that $D_1$ is missed, it will sacrifice B to avoid missing $D_2$. As shown in Line 3, $D_2$ is not missed, then the following C could run normally. If $D_2$ is still missed as in Line 4, the next module C will be sacrificed too.}
\label{fig:pipeline}
\end{figure}

\textbf{Guide the running of subsequent modules:} If the current running state is poor ($\theta t_n > D_n$), the run time of subsequent modules should be sacrificed to avoid missing the e2e deadline. Fig.~\ref{fig:pipeline} visually illustrates this idea of \emph{trade-off}. Line \emph{1} shows a successful pipeline under normal circumstances, with each module completing its task before its respective deadline despite some performance jitter. Module A in the last three lines both encounter performance issues and trigger the alarm. As can be seen from the figure, by sacrificing the next module after timeout occurs, miss of e2e deadline can be avoided as much as possible.

There are two main approaches to sacrificing the run time depending on the task type of the next module: 1. Computation task. The controlling of computation tasks is mainly done by reducing the quality of the results in exchange for less time consuming. Taking algorithms of machine learning as an example, many models \cite{ren2015faster,liu2016ssd,tan2020efficientdet} provide users with different configurations. The better the performance, the slower the model tends to run. 2. The controlling of communication tasks becomes more complicated because of the introduction of the multi-network environment. Here, we also takes the down-sampling \emph{policy} before as an example. The original \emph{policy} uses two different networks to send the raw data and the sampled data respectively, and will wait for the raw data until $D_n$ on the receiving side. Now, to complete the mission of reducing the transmission delay ordered by the real-time mechanism, the \emph{policy} will have both networks send the sampled data, and it will immediately send the message received from either network to the upper-layer application without waiting.

\subsection{Local Backup and Timeout Handling} Finally, to achieve R4, \emph{INTERNEURON} provides a local running backup and timeout handling mechanism. As shown in Fig.~\ref{fig:arch}, a local backup loading will take place at the same time as computation offloading. The local backup is generally executed at the minimum processing quality level to save resource. If an e2e timeout occurs, the local result is returned to the application to keep the system running properly. 

Besides the backup, timeout handling would ask \emph{INTERNEURON} to take active measures to accelerate the next round of operation. First, \emph{policy} will be forced to transmit only the sampled data in the next round. Second, \emph{INTERNEURON} will drastically reduce the value of X: $X=max(X-1,-1)$, so that the next round would impose more stringent delay requirements and the modules would be more likely to sacrifice quality in exchange for speed. Thus, the e2e deadline would not be missed.
\section{\textbf{Evaluation Methodology and Results}}
\label{evaluation}

In this section, we present the experimental methodology and results to demonstrate the effectiveness and efficiency of \emph{INTERNEURON}.

\subsection{Experimental Setup}
The experiments are conducted between two devices, the SoR consists of one Intel Core i7 12700F CPU, 128 GB DDR4 RAM, Nvidia 3070 GPU, and the SoV is equipped with an Nvidia Jetson Nano with 4 core ARM A57, 4 GB RAM. The SoV is connected to the SoR via two wireless NICs. 

\subsection{Dynamic Adjustment of Transmission Time}

\begin{figure}[htb]
\centering
\includegraphics[width=1\linewidth]{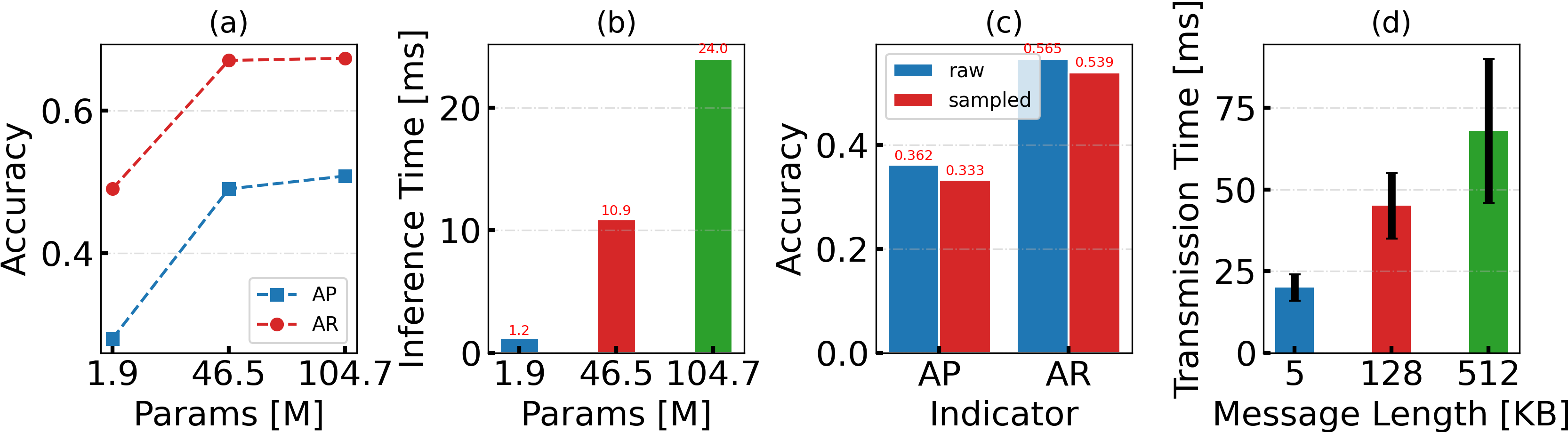}
\caption{(a) shows the quality of inference result using model weights with different number of parameters(YoloV5n has 1.9M, YoloV5l has 46.5M, YoloV5x6 has 104.7M). (b) shows the inference times of these weights. (c) shows the quality of inference result using images with different resolutions(raw image size: 512KB, sampled image size: 128KB) in YoloV5. The performance indicators are Average Precision(AP) and Average Recall(AR), they are measured using YoloV5s with operating parameters: IoU=0.5:0.95, area=all, maxDets=100. (d) shows the transmission time of messages with different length, 5KB message is for transferring inference results. }
\label{fig:adjust}
\end{figure}

First, adjusting computation task is by using different model weights provided by YoloV5 to do the inference. As shown in Fig.~\ref{fig:adjust} (a), with the number of parameters increases, detection accuracy increases accordingly. At the same time, the cost of achieving higher precision is an increase in inference time, as shown in Fig.~\ref{fig:adjust} (b). Based on this observation, a key knob for \emph{INTERNEURON} to adjust is the model size, such that we can tradeoff accuracy for shorter inference time. Second, adjusting transmission delay is by scaling the message length. As shown in Fig.~\ref{fig:adjust} (c) \& (d), although using down-sampled images has a 4\% to 8\% loss in accuracy, the pressure of transferring down-sampled image is much lower. However, since Yolo would eventually scale different lengths of images uniformly into a fixed size before inference, the execution time is not affected by the length of message.

\subsection{Real-Time Guarantee}

\begin{figure}[htb]
\centering
\includegraphics[width=1\linewidth]{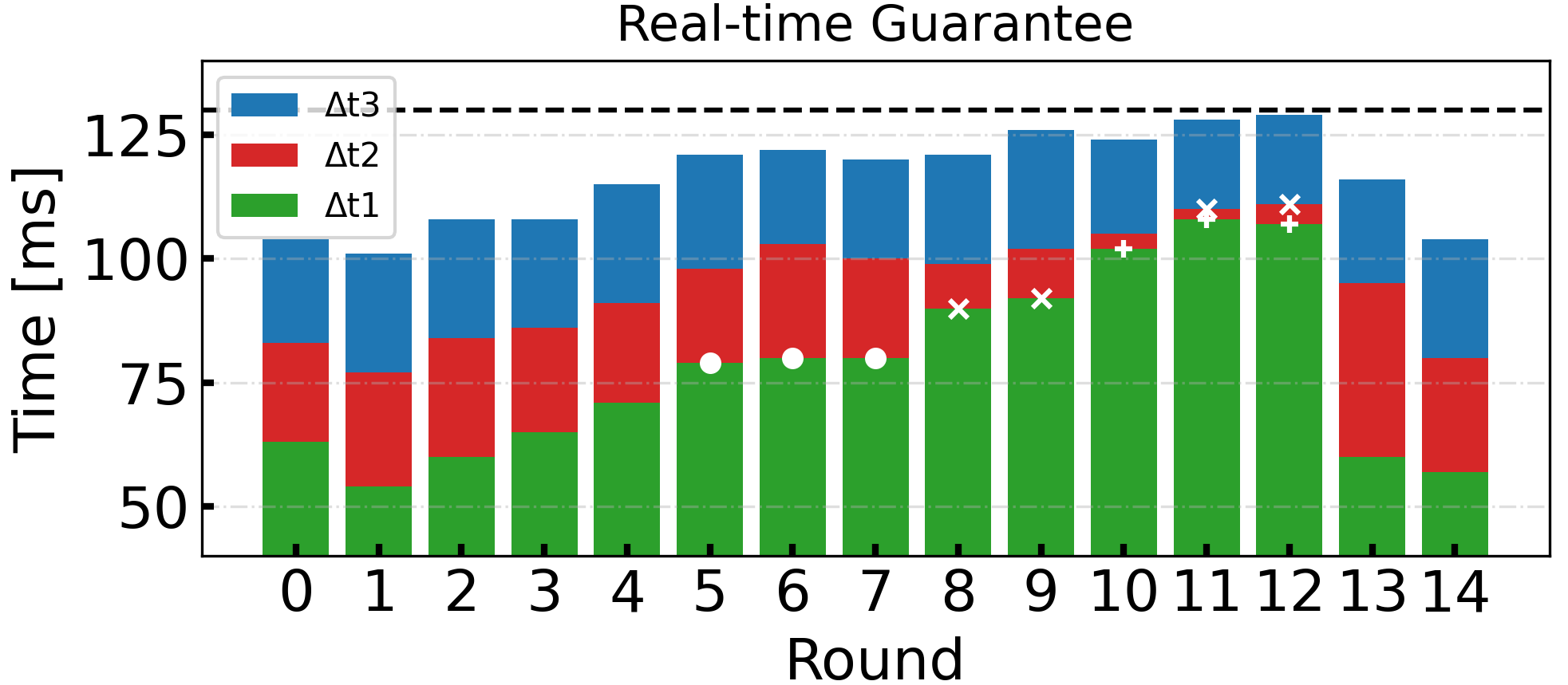}
\caption{We vary the transmission time of images($\triangle t_1$) and show how \emph{INTERNEURON} respond to the changes and avoid a timeout. The e2e timeout time is set to 130ms (marked by the black dashed line), and the parameter \emph{k} for updating $\theta t^{base}_n$ is set to 0.3. White dot in the figure marks the positions where $\theta t_1 = D_1$, 'x' marks the positions where $\theta t_1 > D_1$ or $\theta t_2 > D_2$, and '+' marks the positions where  $\theta t_1$ is much greater than $D_1$.}
\label{fig:rt_guarantee}
\end{figure}

In this section, we increase image transmission delay to stress test \emph{INTERNEURON}'s capability to provide real-time guarantee, a typical process is shown in Fig.~\ref{fig:rt_guarantee}. In the first three rounds, the system is in a normal state, where all the modules are outputting the highest quality results. Then we start to increase the raw image transmission delay. In rounds 3 \& 4, $\theta t_1$ is still within $D_1$, so all tasks continue to run normally. However, with the reference e2e run time ($\theta t^{base}_3$) increases (X decreases), \emph{INTERNEURON} becomes more sensitive to delay jitters. Next, in rounds 5-7, raw image transmission time exceeds $D_1$. According to the down-sampling \emph{policy} introduced in Section \ref{design}, before using the sampled data, \emph{INTERNEURON} would wait for the raw image until $D_1$. Therefore, at the white dots in the figure, although the system ultimately uses the sampled data, the image transmission time is exactly equal to $D_1$.

Then, we increase the latency of the other network to make the delay of transmitting sampled data exceeds $D_1$ in rounds 8-12. As a result, $D_1$ is missed, and \emph{INTERNEURON} has to sacrifice the computation module to reduce time consumption. The computation module will select the appropriate weights according to the timeout degree of the transmission delay. In rounds 10-12, the computation module selects the weight with fewer parameters than rounds 8-9 because of the serious timeout of the transmission task. And in round 11 \& 12, $D_2$ is missed, so the third module is sacrificed too.

\subsection{Timeout Handler}
Experimental results show that the continuous timeouts can be effectively eliminated by adjusting X to dynamically limit the execution quality in the next round (frequency of two consecutive timeouts has been reduced by 78\%). Consider a possible X jitter problem: After round i misses the deadline, the timeout handler adjusts X to a negative value, and round i+1 finishes on time. But because round i+1 takes so little time (caused by drastically reduced quality), X will become too large again. And then round i+2 will run at a high quality, which is easy to cause timeout again since the environment is likely to be unstable. Our solution is: when X is negative, completing the pipeline on time in each round will add 0.2 to X, until X is positive. Experimental results verify that this method can effectively eliminate the occurrence of X jitter.

\begin{table}[h]
\caption{INTERNEURON \& ERDOS.}
\label{tab:compare}
\resizebox{\textwidth/2}{!}{
\begin{threeparttable}[b]
\begin{tabular}{ccccc} 
\hline
            & e2e Timeout & * Average e2e run time & Average AP & Average AR\\ \hline
INTERNEURON & 8 rounds & 114msec & 0.472 & 0.624\\
ERDOS       & 152 rounds & 109msec & 0.508 & 0.673\\ \hline
\end{tabular}
\begin{tablenotes}
\footnotesize
\item[*] The average e2e run time is calculated without counting the rounds where timeout occurs.
\end{tablenotes}
\end{threeparttable}
}
\end{table}

\subsection{Reduction of Timeouts}
Finally, we compare \emph{INTERNEURON} against ERDOS, which has been introduced in Section \ref{background}. We evaluated both frameworks by executing the pipeline shown in Fig.~\ref{fig:typical_pipeline} for 300 rounds. During the experiments, we randomly injected delays into computation or communication modules. As shown in Table~\ref{tab:compare}, the experimental results confirm that the number of timeouts in \emph{INTERNEURON} is 95\% fewer compared to ERDOS, thus verifying \emph{INTERNEURON}'s capability to greatly improve communication reliability in IVCAD. 

\section{\textbf{conclusion}}
\label{conclusion}
This paper proposes INTERNEURON, a middleware to achieve high communication reliability in IVCAD. With the aim of achieving optimal IVCAD reliability, INTERNEURON exploits the characteristics of existing multi-network communication environment to dynamically match IVCAD applications to the underlying communication networks. INTERNEURON is designed to be lightweight, high performance, reliable, and easy-to-deploy.  Evaluation results have confirmed the effectiveness of INTERNEURON, which reduces deadline violations by more than 95\%. As a next step, we plan to extend INTERNEURON to more usage scenarios beyond IVCAD, including vehicle-to-vehicle communication, and cooperative robots.

\clearpage
\bibliographystyle{IEEEtran}
\bibliography{ref}

\vspace{12pt}
\end{document}